\documentclass[11pt,a4paper]{article}
\usepackage{times,latexsym}
\usepackage{url}
\usepackage[T1]{fontenc}

\usepackage{microtype}
\usepackage{graphicx}
\usepackage{caption}
\usepackage{subcaption}
\usepackage{amsmath}
\usepackage{amssymb}
\usepackage[normalem]{ulem}
\usepackage{float}
\usepackage{multirow}
\usepackage[acceptedWithA]{tacl2021v1}
\usepackage{xspace,mfirstuc,tabulary}

\newif\iftaclinstructions
\taclinstructionsfalse % AUTHORS: do NOT set this to true
\iftaclinstructions
\renewcommand{\confidential}{}
\renewcommand{\anonsubtext}{(No author info supplied here, for consistency with
TACL-submission anonymization requirements)}
\newcommand{\instr}
\fi

\iftaclpubformat % this "if" is set by the choice of options

\else

\fi

\newcommand{\SHORTENED}[1]{} % stuff that can be restored for camera-ready

\newcommand{\gap}{\vspace{2mm}}
\newcommand{\framework}{NeLLCom}

\title{Communication Drives the Emergence of Language Universals in \\ Neural Agents: Evidence from the Word-order/Case-marking Trade-off}

\author{
  Yuchen Lian$^\diamond$ $^\dagger$
  \qquad
  Arianna Bisazza$^\ddagger$
  \qquad
  Tessa Verhoef$^\dagger$
  \\
  \ \\
  $^\diamond$Faculty of Electronic and Information Engineering, Xi'an Jiaotong University
  \\
  $^\dagger$Leiden Institute of Advanced Computer Science, Leiden University
  \\
  \texttt{\{y.lian, t.verhoef\}@liacs.leidenuniv.nl}
  \\
  $^\ddagger$Center for Language and Cognition, University of Groningen \\
  \texttt{a.bisazza@rug.nl}
}

\date{}

\begin{document}
\maketitle
\begin{abstract}
Artificial learners often behave differently from human learners in the context of neural agent-based simulations of language emergence and change.
A common explanation is the lack of appropriate cognitive biases in these learners.
However, it has also been proposed that more naturalistic settings of language learning and use could lead to more human-like results. 
We investigate this latter account focusing on the word-order/case-marking trade-off, a widely attested language universal that has proven particularly hard to simulate. % \cite{chaabouni-etal-2019-word,lian-etal-2021-effect}. 
%inspired by human learner experiments in \cite{fedzechkina2017balancing} where miniature languages with different proportions of order and case marking are designed to be learned.
We propose a new Neural-agent Language Learning and
Communication framework \mbox{(NeLLCom)} where pairs of speaking and listening agents first learn a miniature language via supervised learning, and then optimize it for communication via reinforcement learning.
Following closely the setup of earlier human experiments, %\cite{fedzechkina2017balancing}, 
we succeed in replicating the trade-off with the  new framework without hard-coding specific biases in the agents. We see this as an essential step towards the investigation of language universals with neural learners. 
%Although some of our results differ from those of the human experiments,  
%we make a potentially crucial step towards developing a neural-agent framework that can replicate an important pattern of natural languages without the need to ‘hard-code’ any learning bias. 
\end{abstract}

\section{Introduction}

The success of deep learning methods for natural language processing has triggered a renewed interest in agent-based computational modeling of language emergence and evolution processes \citep{lazaridou2020emergent, chaabouni2022emergent}. %boer2006computer,steels2016agent
An important challenge in this line of work, however, is that such artificial learners often behave differently from human learners \citep{galke2022emergent,rita2022emergent,chaabouni-2019-antiefficient}.

One of the proposed explanations for these mismatches is the difference in cognitive biases between human and neural-network (NN) based learners. For instance, the neural-agent iterated learning simulations of  \citet{chaabouni-etal-2019-word} and \citet{lian-etal-2021-effect} did not succeed in replicating the trade-off between word-order and case marking, which is widely attested in human languages \cite{sinnemaki2008complexity,futrell-etal-2015-quantifying} and has also been observed in miniature language learning experiments with human subjects \cite{fedzechkina2017balancing}. Instead, those simulations resulted in the preservation of languages with redundant coding mechanisms, which the authors mainly attributed to the lack of a human-like least-effort bias in the neural agents.
Besides human-like cognitive biases, it has been proposed that more natural settings of language learning and use could lead to more human-like patterns of language emergence and change \citep{mordatch2018emergence,lazaridou2020emergent,kouwenhoven2022emerging,galke2022emergent}. 
In this work, we follow up on this second account and investigate whether neural agents that \textit{strive to be understood} by other agents display more human-like language preferences.

% === AB The following paragraph is very redundant with the  following one! ==
%Keeping the focus on the word-order/case-marking trade-off, we show that the right combination of a \textit{supervised language learning} protocol with a \textit{communication task} between agents can lead to results that closely resemble those in human experiments, without the need to implement complex human-like biases (such as least-effort) into the artificial learners.

To achieve that, we design a Neural-agent Language Learning and Communication (NeLLCom) framework that combines Supervised Learning (SL) with Reinforcement Learning (RL), inspired by \citet{lazaridou-etal-2020-multi} and \citet{lowe2019interaction}.
Specifically, we use SL to teach our agents predefined languages characterized by different levels of word order freedom and case marking. %, inspired by the artificial languages used by \citet{fedzechkina2017balancing} in experiments with human participants.
Then, we employ RL to let pairs of speaking and listening agents talk to each other while optimizing communication success (also known as \textit{self-play} in the emergent communication literature).

We closely compare the results of our simulation to those of an experiment with a very similar setup and miniature languages involving human learners \citep{fedzechkina2017balancing}, and show that a human-like trade-off can indeed appear during neural-agent communication.
Although some of our results differ from those of the human experiments, we make an important contribution towards developing a neural-agent framework that can replicate language universals without the need to hard-code any ad-hoc bias in the agents. 
We release the NeLLCom framework\footnote{All code and data are available at \url{https://github.com/Yuchen-Lian/NeLLCom}} to facilitate future work simulating the emergence of different language universals.

\section{Background}

\paragraph{Word order \textit{vs.} case marking trade-off}
A research focus of linguistic typology is to identify \textit{language universals} \cite{greenberg1963universals}, i.e. patterns occurring systematically among the large diversity of natural languages. The origins of such universals are object of long-standing debates.
\SHORTENED{add some citations here: Lupyan's niche hypothesis, Levshina..?}
The trade-off between word order and case marking is an important and well-known example of such a pattern that has been widely attested \cite{comrie1989language,blake2001case}.
% \citet{greenberg1963universals} noted a very strong correlation between dominant S-O-V (subject object verb) word order and case marking. No such correlation exists for S-V-O (subject verb object) languages.
Specifically, languages with more flexible constituent order tend to have rich morphological case systems (e.g. Russian, Tamil, Turkish), while languages with more fixed order tend to have little or no case marking (e.g. English or Chinese).
%This inverse correlation has been widely explored in the field of linguistic typology. 
Additionally, quantitative measures also
%in \citet{sinnemaki2008complexity} and \citet{futrell-etal-2015-quantifying} 
revealed that the functional use of word order has a statistically significant inverse correlation with the presence of morphological cases based on typological data \cite{sinnemaki2008complexity,futrell-etal-2015-quantifying}.
% with a sample of 50 languages.
% \AB{is the sample of 50 languages also used in Sinnemaki or only Futrell? only in Sinnemaki}
% Although most theories have argued that these common design features are shaped by human cognitive constraints and pressures during communication and transmission, there are still debates about whether these biases are specific to language. 

Various experiments with human participants  \cite{fedzechkina2012language,fedzechkina2017balancing,tal2022redundancy} were conducted to reveal the underlying cause of this correlation.
In particular, \citet{fedzechkina2017balancing}, who highly inspired this work, 
applied a miniature language learning approach to study whether the trade-off could be explained by a human learning bias to reduce production effort while remaining informative.
In their experiment, two groups of 20
participants were asked to learn one of two predefined miniature languages. 
Both languages contained optional markers but differed in terms of word order (fixed \textit{vs.} flexible). 
After three days of training, both groups reproduced the initial word order distribution, however the flexible-order language learners used case marking significantly more often than the fixed-order language learners. Moreover, an asymmetric marker-using strategy was found in the flexible-order language learners, %where utterances with the cross-linguistically less frequent order had markers more often.
whereby markers tended to be used more often in combination with the less frequent language.
Thus, most participants displayed an inverse correlation between the use of constituent order and case marking \textit{during} language learning, which the authors attributed to a unifying information-theoretic principle of balancing effort with robust information transmission.
%provide evidence for learners’ preference to trade off efficiency with robustness in information transmission  to some extent, working as one of the main factors in shaping human languages.

% \old{Glancing at Human results}
% \YLrevise{
% To give a better expectation for the following neural agents\' performance, here we generalize the main learning results from human subjects conducted by \citet{fedzechkina2017balancing}. 
% In their work, 40 participants were separated into two groups and were presenting one of the languages described in Section.~\ref{sec:grammars}. They were informed to learn this novel language by watching short videos and hearing the corresponding descriptions. 
% For ordering preferences, both groups of learners maintained the original order proportions.  
% For case marking preferences, flexible-order language learners use case marking significantly more often compared to fixed-order language learners. 
% An asymmetric marker-using strategy was also found in the flexible-order language learning results, where utterances with the cross-linguistically less frequent order have markers more often.
% Besides these statistical analyses, 
% a trade-off between robust information transmission and effort is guided as a unifying information-theoretic principle for all learners, with an overall strong bias to reduce uncertainty \cite{fedzechkina2017balancing}.
%  }

\paragraph{Agent-based simulations of language evolution}
%Prior to the recently renewed interest in agent-based simulations of language emergence, 
Computational models have been used widely to study the origins  of language structure \cite{kirby2001spontaneous, steels2016agent, boer2006computer,van2003language}. In  particular,
\citet{lupyan2002case} were able to mimic the human acquisition patterns of four languages with very different word order and case marking properties, using a simple recurrent network \cite{elman1990finding}.
%studied agent acquisition of languages with different case marking and word order patterns with sequential learning devices and demonstrated that after training, a simple Elman network (SRN) \cite{elman1990finding} is able to mimic the syntactic performance of child language acquisition, although certain limitations may apply.

%With modern deep learning methods, a growing number of recent studies show a return of this computational modeling approach to reconstruct natural language patterns with neural agents
Modern deep learning methods have also been used to simulate patterns of language emergence and change \cite{chaabouni-2019-antiefficient,chaabouni-etal-2019-word,chaabouni-etal-2020-compositionality,chaabouni2021communicating,lian-etal-2021-effect,lazaridou2018emergence,ren2020compositional}. 
Despite several interesting results, many report the emergence of languages and patterns that significantly differ from human ones.
For example, \citet{chaabouni-2019-antiefficient} found an anti-efficient encoding scheme that surprisingly opposes Zipf's Law, a fundamental feature of human language.
\citet{rita-etal-2020-lazimpa} obtained a more efficient encoding by explicitly imposing a length penalty on speakers and pushing listeners to guess the intended meaning as early as possible.
% SHORTENED{Compositionality, as another intriguing feature, arises with the deliberate design of the networks or mostly does not appear in the agents' emergent communication protocols \cite{lazaridou2018emergence,gupta-etal-2020-compositionality,baroni2020linguistic}.}
%\ABlater{address Tessa's comment on compositionality}
\SHORTENED{newline}
Focusing on the order/marking trade-off,
\citet{chaabouni-etal-2019-word} implemented an iterated learning setup inspired by % \{citet{kirby2001spontaneous, kirby2014iterated}, 
\citet{kirby2014iterated}
where agents acquire a language through SL, and then transmit it to a new learner, iterating over multiple generations. The trade-off did not appear in their simulations.
\newcite{lian-etal-2021-effect} extended the study %of \citet{chaabouni-etal-2019-word} 
by introducing several crucial factors from the language evolution field (e.g. input variability, learning bottleneck), but no clear trade-off was found.
% \newcite{lian-etal-2021-effect} recently explored the emergence of this trade-off with the iterated learning framework in which agents acquired a language through supervised learning, and then transmitted this to a new learner, which was repeated over several generations. Crucial factors known from studies in language evolution (e.g. a bias for efficiency or a learning bottleneck) were introduced into the agent setups, but no clear trade-off was found.
To our knowledge, no study with neural agents has successfully replicated the emergence of this trade-off so far.

%\subsection{Deep Multi-Agent Systems for Emergent Communication}
%5. We follow hypo2, design new framework inspired by (lowe, lazaridou)
%Different goals : getting agents to speak to each other, no predefined language, more practical , not intersted in specific patterns of language change

%\subsection{\citet{fedzechkina2017balancing}: Miniature Language Learning}

\section{\framework: Language Learning and Communication Framework}
\label{sec:framework}

\begin{figure*}
\begin{subfigure}{.3\textwidth}
  \centering
  \includegraphics[width=\columnwidth]{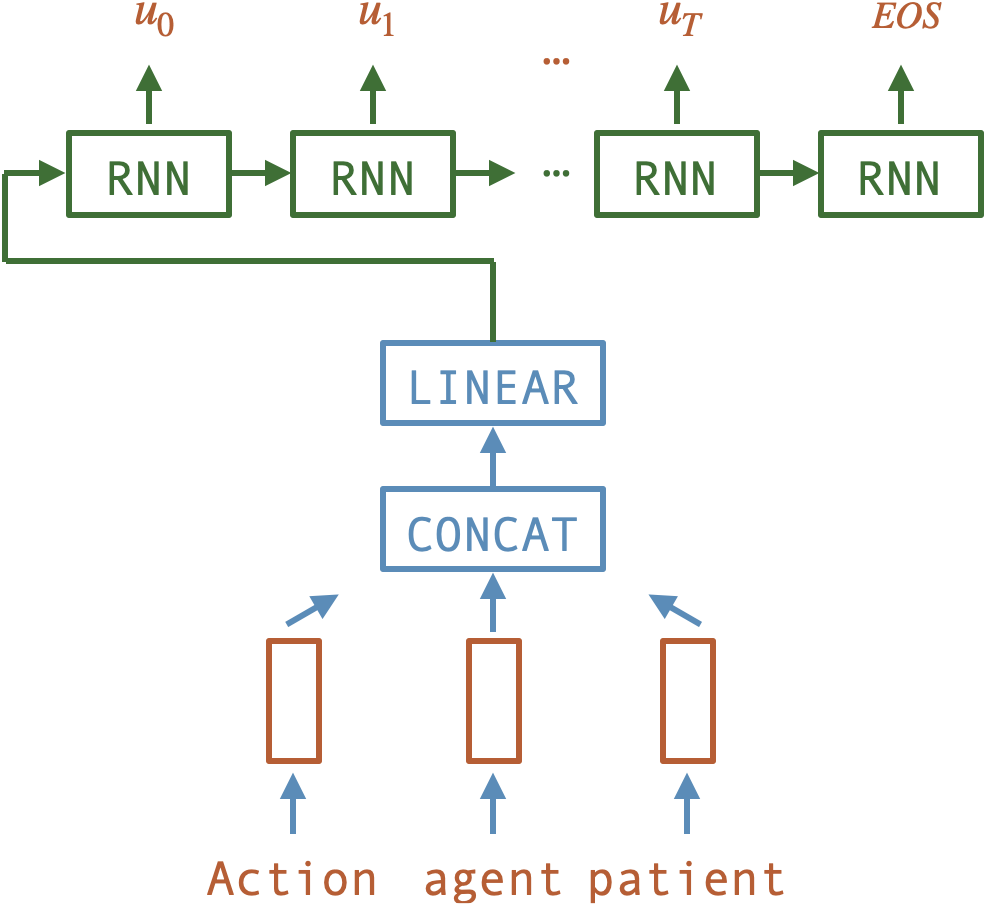}
  \caption{Speaking Agent}
  \label{fig:spker}
\end{subfigure}%
\begin{subfigure}{.3\textwidth}
  \centering
  \includegraphics[width=\columnwidth]{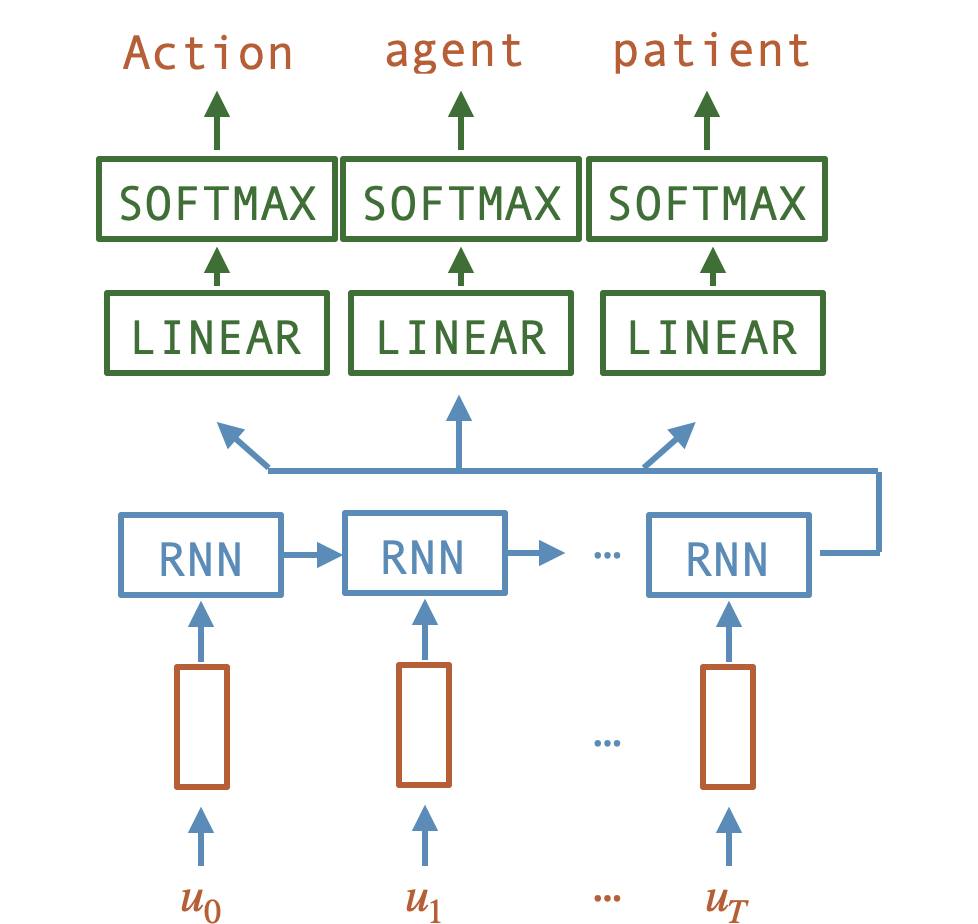}
  \caption{Listening Agent}
  \label{fig:lster}
\end{subfigure}%
\begin{subfigure}{.4\textwidth}
  \centering
  \includegraphics[width=\columnwidth]{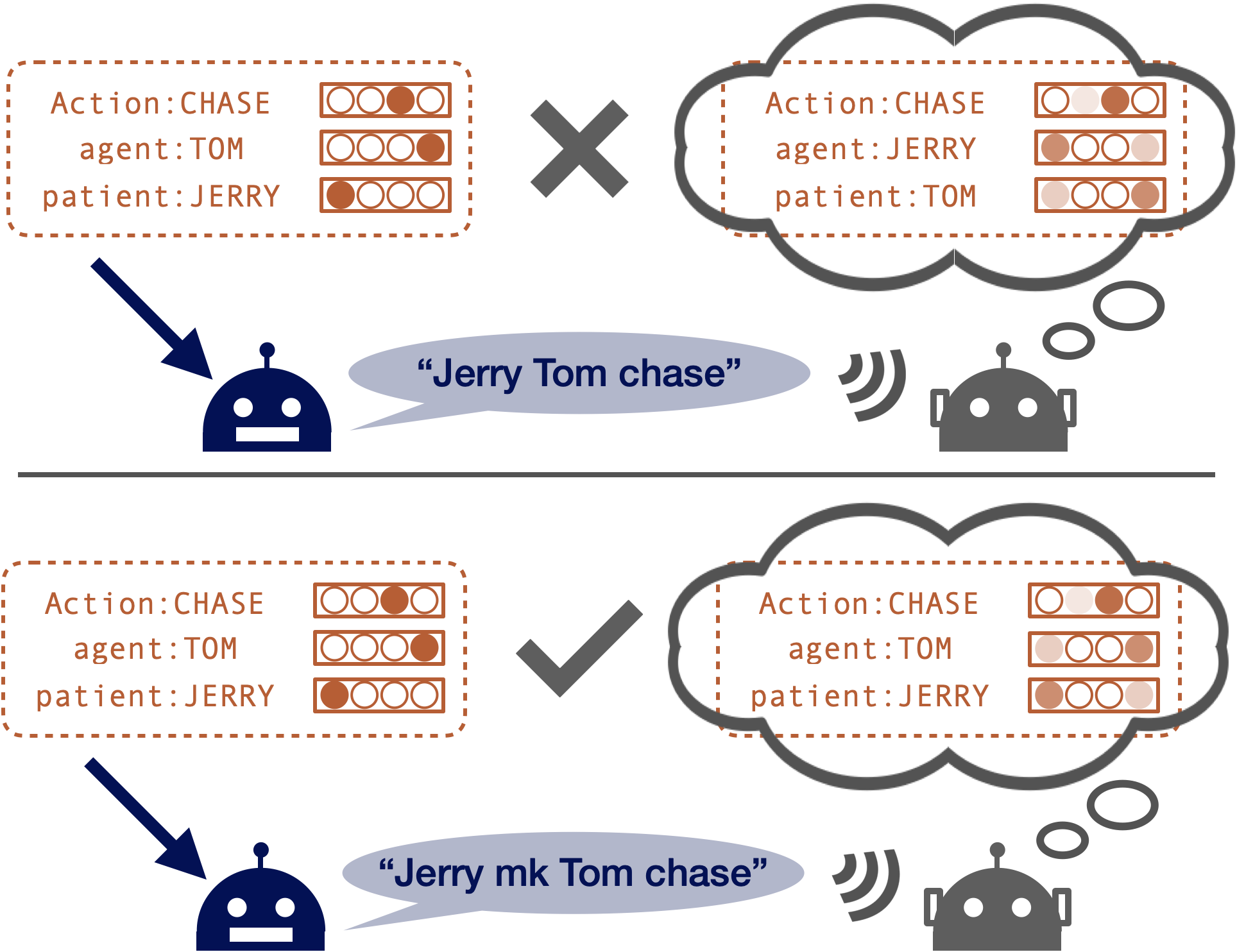}
  \caption{Agents Communication}
  \label{fig:comm}
\end{subfigure}%
\caption{Agents architecture and a high-level overview of the meaning reconstruction game.}
\label{fig:fm}
\end{figure*}

This section introduces the Neural-agent Language Learning and Communication (\textbf{\framework}) framework, which we make publicly available.
%starting from the agent architectures followed by the different training procedures and their combination. 
Our goal differs from that of most work in emergent communication, where language-like protocols are expected to \textbf{arise from sets of random symbols} through interaction \cite{lazaridou2018emergence,havrylov2017emergence,chaabouni-2019-antiefficient,chaabouni2022emergent,bouchacourt-baroni-2018-agents}.
We are instead interested in observing how \textbf{a given language with specific properties changes} as the result of learning and use.
Specifically, in this work, agents need to learn miniature languages with varying word order distributions and case marking rules.
While this can be achieved by a standard SL procedure, we hypothesize that \textbf{human-like regularization patterns} will only appear when our agents strive to be understood by other agents.
We simulate such a need via RL, using a measure of communication success as the optimization objective.

Similar SL+RL paradigms have been used in the context of communicative AI \cite{li-etal-2016-deep,strub2017end,das2017learning}. %,lu2020countering}. 
In particular, \citet{lazaridou-etal-2020-multi} and \citet{lowe2019interaction} explore different ways of combining SL and RL to teach agents to communicate with humans in natural language.
A well-known problem in that setup is that languages tend to \textit{drift} away from their original form as agents adapt to communication. 
% THESIS {With the mission of grounding in natural language, this language drifting phenomenon can be detrimental when it starts to diverge too much from English, and several solutions for it have been proposed \citep{lu2020countering,lee-etal-2019-countering,lazaridou-etal-2020-multi}. In our case, we do not want to prevent the drift. Instead, we are actually interested in observing it and comparing it to the patterns observed in human experiments.}
In our context, we are specifically interested in studying how this drift compares to human experiments of artificial language learning.
Our implementation is partly based on the EGG toolkit\footnote{\url{https://github.com/facebookresearch/EGG}} \cite{kharitonov-etal-2019-egg}.

\subsection{The Task}
\label{sec:task}

% \paragraph{The Task}
NeLLCom agents communicate about a simplified world using pre-defined artificial languages. 
Speaking agents convey a meaning $m$ by generating an utterance $u$, whereas listening agents try to map an utterance $u$ to its respective meaning $m$. 
The meaning space includes agent-patient-action triplets, such as \textit{dog-cat-follow, dog-mouse-follow}, defined as triplets $m=\{A,a,p\}$, where $A$ is an action, $a$ the agent, and $p$ the patient.
%(cf. Section.~\ref{sec:meaning-space}).
Utterances are variable-length sequences of symbols taken from a fixed-size vocabulary: $u=[w^1, ..., w^I]$, $w^i \in V$. 
Evaluation is conducted on meanings unseen during training.

\subsection{Agent Architectures}

Both speaking and listening agents contain an encoder and a decoder, however their architectures are mirrored as the meanings and sentences are represented differently (see Fig.~\ref{fig:fm}).

\paragraph{Speaker: linear-to-sequence}
In a speaker network ($S$), the encoder receives the hot-vector representations of $A$, $a$, and $p$, and projects them to latent representations or embeddings. The order of these three elements is irrelevant.
The concatenation of the embeddings followed by a linear layer becomes the latent meaning representation,\footnote{We opted for simple architectural choices whenever possible.  Adding a non-linearity to the meaning encoder did not affect the results.} 
based on which the Recurrent Neural Network (RNN) decoder generates a sequence of symbols.\footnote{We implement the speaker's decoder and the listener's encoder as one-layer Gated Recurrent Units (GRU) \cite{chung2014empirical} following previous work on language emergence \cite{dessi2019focus,chaabouni-etal-2020-compositionality}. The latter paper, in particular, reports slower convergence with LSTM than GRU, and a lack of success at adapting Transformers to their setup.}

\paragraph{Listener: sequence-to-linear}
The listener network ($L$) works in the reverse way: its RNN encoder takes an utterance as input and sends its encoded representation to the decoder, which tries to predict the corresponding meaning.
Specifically, the final RNN cell is fed to the decoder, which passes it through three parallel linear layers, for $A$, $a$, and $p$, respectively. Finally, each of the three elements is generated by a softmax layer.

\gap

Unlike the agents of \citet{chaabouni-etal-2019-word} and \citet{lian-etal-2021-effect}, our agents can only behave as either speaker or listener, but not both. %Due to the different nature of meaning and utterance in our setup, this is not straightforward to achieve. 
\citet{chaabouni-etal-2019-word} achieved this by tying input and output embeddings, however they reported only a minor effect on the results.
As another difference, we represent meanings as unordered attribute-values instead of sequences, which we find important to avoid any ordering bias in the meaning representation.
\SHORTENED{
In addition, we also change the structural design. 
Instead of processing both meaning and utterance sequentially, i.e. Seq2Seq structure in \citet{chaabouni-etal-2019-word} and \citet{lian-etal-2021-effect}, 
we opt for linear processors for the meaning and sequential processors for the utterance. 
Thus we can treat the meanings as unordered sets of value-attributes to avoid any bias towards the order of action, agent, patient in the meaning vector.
}
% A possible solution can be to link the networks with a common latent space as shared knowledge about the language.
% While this work focuses on how the form of communication affects agents language producing, we are not planning to complicate our agent structures. We leave the problem of how to implement this as our further work.  Moreover, there is still debating about how human processing linguistic knowledge. Human speaking and listening ability can be controlled in the separate functional area in brain.
We note that the framework is rather general: in future studies, it could be adapted to different meaning spaces and different artificial languages, as well as different types of neural sequence encoders/decoders.

\subsection{Supervised Language Learning}
\label{sec:sv}

SL is a natural choice to teach agents a specific language.
This procedure requires a dataset $D$ of meaning-utterance pairs $\langle m,u \rangle$ where $u$ is the gold-standard generated for $m$ by a predefined grammar (see grammar details in Section.~\ref{sec:grammars}).
The learning objectives differ between speaker and listener agents.

\paragraph{Speaker}
Given $D$, speaker's parameters $\theta_S$ are optimized by minimizing the cross-entropy loss:
\begin{equation}
Loss^{sup}_{(S)}  = -\sum_{i=1}^{I} \log p_{\theta_S}(w^i|w^{<i},m)
\end{equation}

\noindent 
where $w^i$ is the $i^{th}$ word of the gold-standard utterance $u$.
Notice that SL implies a teacher forcing procedure \citep{goodfellow2016deep}, meaning that at each timestep the gold history $w^{<i}$ is used to predict the next word $w^i$ and update the network weights accordingly.

\paragraph{Listener}
Given $D$, listener's parameters $\theta_L$ are optimized by minimizing the cross-entropy loss:
\begin{multline}
Loss^{sup}_{(L)} = 
- (\log p_{\theta_L}(a | u) \\
+ 
\log p_{\theta_L}(p | u) + \log p_{\theta_L}(A | u))
\label{fuc:lst_loss}
\end{multline}

\begin{table*}[h!t]
\begin{center} \small
\begin{tabular}{c|c|l|c|c}
\hline \bf language & \bf word order & \bf case marking & \bf $m=\{A,a,p\}$ & \bf $u$ \\ 
\hline
% fix+full & 100\% SOV & 100\% on OBJ \\
fix+op  & 100\% SOV & 66.7\% on OBJ & \textsc{chase tom jerry} & \it Tom Jerry chase $|$ Tom Jerry mk chase \\
\hline
% flex+full & 50\% SOV, 50\% OSV & 100\% on OBJ \\
\multirow{2}{*}{flex+op} & 50\% SOV, & \multirow{2}{*}{66.7\% on OBJ} & \multirow{2}{*}{\textsc{chase tom jerry}} & \it Tom Jerry chase $|$ Tom Jerry mk chase $|$ \\
 & 50\% OSV &  &  & \it Jerry Tom chase $|$ Jerry mk Tom chase\\
\hline
\end{tabular}
\end{center}
\caption{\label{tab:language-table} The two miniature grammars used in this study, along with meaning-utterance $\langle m,u \rangle$ examples.}
\end{table*}

%\subsection{Learning to Communicate}
\subsection{Optimizing Communication Success}
\label{sec:rl}
While SL may be sufficient to (perfectly) learn a given meaning-to-signal mapping and vice versa, we are interested in whether and how such language changes as a result of repeated usage.
Following a long-established practice of simulating emergent communication with humans and computer agents in language evolution 
\citep{steels1997synthetic, steels2016agent, selten2007emergence,galantucci2011experimental}, and more recently also in the computational linguistics literature 
\citep{bouchacourt-baroni-2018-agents,lazaridou2018emergence,lazaridou-etal-2020-multi,lowe2019interaction,havrylov2017emergence,evtimova2018emergent},
we simulate communication with a meaning reconstruction game where a speaker $S$ learns to convey meanings $m$ to a listener $L$ using utterances $\hat{u}$ in the language it has learned by SL. 
The goal for both agents is to maximize a shared reward evaluated by the listener's prediction.
%\THESIS{
%The structure of this pipeline looks similar to an Auto-encoder except that $u_{spk}$ generated by the speaker through sampling from the output distribution which is not differentiable.}
For this phase, we adopt the classical policy-based algorithm REINFORCE \cite{williams1992simple}.
Specifically, we optimize:
\begin{equation}
    Loss_{(S,L)}^{comm} = -r^L(m, \hat{u})*\sum_{i=1}^{I} \log p_{\theta_S}(w^i|w^{<i},m)
\end{equation}
where 
% $\hat{m_{L}}(\hat{u})$ is the listener's prediction given $\hat{u}$ and 
$r^L(m, \hat{u})$ is defined as 
the cross-entropy loss
% the similarity 
between input meaning $m$ and listener's prediction: % in Equation \ref{fuc:comm_r}.
\begin{equation}
r^L(m, \hat{u}) = \sum_{e\ \in\ m=\{a, p, A\}} \log p_{\theta_L}(e | \hat{u})
\label{fuc:comm_r}
\end{equation}

\subsection{Combining Supervision and Communication}
%Combining SL and RL into a single training procedure would be ideal to simulate a human learner with a need to be informative, but is technically difficult to achieve. 
%
We adopt the simplest possible way of combining SL and RL, which is to first train the agents by SL until convergence and then continue training them by RL to maximize the communicative reward.\footnote{This procedure corresponds to \textit{reward fine-tuning} in \citet{lazaridou-etal-2020-multi} and to \textit{sup2sp} in \citet{lowe2019interaction}.}
While more sophisticated combination techniques were proposed recently \cite{lazaridou-etal-2020-multi,lowe2019interaction}, we find this simple SL+RL sequence to work well in our context, and leave an exploration of other techniques to future work.

Crucially, using communication success as task reward rather than forcing agents to imitate given training pairs $\langle m,u \rangle$ allows agents to depart from the initially learnt grammar, as long as the new language remains understandable by other agents. 
This principle is well studied in the framework of Rational Speech Act (RSA) \cite{goodman2016pragmatic} which implemented utterance understanding from a social cognition aspect.
If a language is suboptimal for an agent, e.g. in terms of efficiency or ambiguity, we expect it to change throughout multiple communication rounds.
Note that the listener's role can also be interpreted as that of a \textit{speaker-internal} monitoring system that predicts the chance of a message to be understood by a listener \textit{before} uttering it \citep{ferreira2019mechanistic}. %\AB{Tessa OK? Add refs?} \TV{Yes that seems intuitive, in cogsci we call this audience design, I added a ref!}

\section{Experimental Setup}

We use NeLLCom to replicate the results of \citet{fedzechkina2017balancing}, who taught human subjects miniature languages with varying order distributions.
Subjects watched short videos of two actors performing simple transitive events (e.g. \textit{a chef hugging a referee}) accompanied by spoken descriptions in the novel language.\footnote{Sentence learning was preceded by a noun learning phase which we do not model in our experiments. For more details on the human training process, see \citet{fedzechkina2017balancing}.}
We adopt the same setup, with two notable differences:
(i) our agents do not take videos or images as input, but triplets of symbols representing agent, patient and action, respectively (see Section.~\ref{sec:framework});
(ii) descriptions are not spoken but written, and words are represented by dummy strings (such as \textit{noun-1}, \textit{verb-2}, etc.) instead of English-like sounding nonce words.
Thus, we abstract away from the problem of (i) mapping visual input to structured meaning representations and (ii) mapping continuous audio signals to discrete word representations, respectively. Dealing with these interfaces is necessary when working with humans, but not with neural agents. Moreover, none of them are a core aspect of our investigation.

\subsection{Miniature Languages}
\label{sec:grammars}

Following \citet{fedzechkina2017balancing}, we consider two head-final languages: 
one with fixed order and optional case markers (\textbf{fix+op}), and one with flexible order and optional case markers (\textbf{flex+op}). 
Optional marking means that 2/3 of all objects are followed by a special mark (the token \textit{mk}), whereas subjects are never marked.
Possible constituent orders are SOV and OSV: the fixed-order language uses always SOV, while the flexible-order one uses both with a probability of 50-50\%.
The two languages are illustrated in Table~\ref{tab:language-table}.

In fix+op, order is informative and sufficient to disambiguate grammatical functions. Case marking is therefore a redundant cue.
In flex+op, order is uninformative therefore marking --when present--  is important to recover the meaning.
The hypothesis that language learning and use create biases towards efficient communication systems \citep{gibson2019efficiency, fedzechkina2012language} yields two predictions: fix+op is expected to become less redundant (by a decrease of case marking) whereas flex+op should become more predictable (by an increase of marking \textit{or} a more consistent order).

\subsection{Meaning Space}
\label{sec:meaning-space}

The meaning space used by \citet{fedzechkina2017balancing} included 6 entities and 4 actions, resulting in a total of 6$\times$(6$-$1)$\times$4$=$120 possible meanings (an entity cannot be agent and patient at the same time).  
While suitable for human learners, such a space is too small to train neural agents \cite{zhao2018bias,chaabouni-etal-2020-compositionality}. % have shown that a sufficiently large sample size is necessary to train deep networks to generalize.  
In preliminary experiments, we found that our learners converge well with a meaning space size of 720 (10 performers and 8 actions in our languages).
\SHORTENED{
This also aligns with the results in \citet{chaabouni-etal-2020-compositionality} where generalization in language only appeared when input size was greater that 625.}

To test the agents' ability to convey new meanings, we split our dataset into 66.7\% training and 33.3\% testing. %i.e. 480 meaning-utterance pairs in the training set and 240 pairs in the test set. 
We also ensure that each entity and action of the meaning space appears at least once in the training set.
To prevent the agents from memorizing spurious correlations between a meaning and a particular order or marking choice, we regenerate a new utterance per meaning (according to the same grammar) after each epoch of SL.

\gap

See Appendix.~\ref{app:trainDetails} for details on the datasets and training process.

\section{Supervised Learning Results}

\label{sec:sv_result}
We start by evaluating the agents' ability to learn to speak or listen in a fully supervised way, that is, using the generated meaning-utterance pairs from a specific language as labeled data. 

\begin{figure}[!ht]
\centering
\captionsetup{singlelinecheck=off}
\begin{subfigure}{.25\textwidth}
  \centering
  \includegraphics[width=\columnwidth]{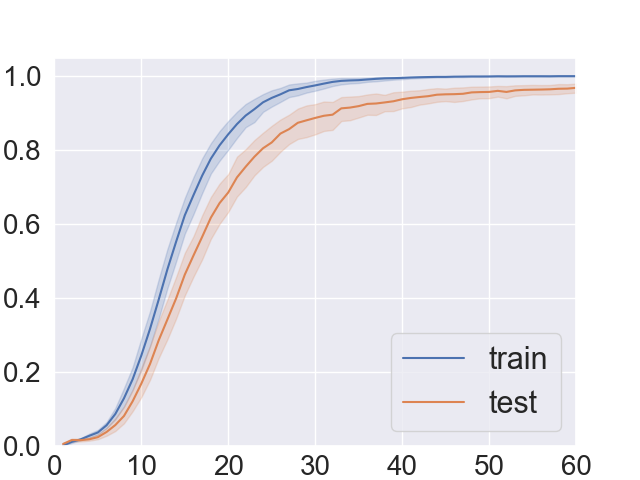}
  \caption{Lst Acc fix+op}
  \label{fig:fix_lst_sv_acc}
\end{subfigure}%
\begin{subfigure}{.25\textwidth}
  \centering
  \includegraphics[width=\columnwidth]{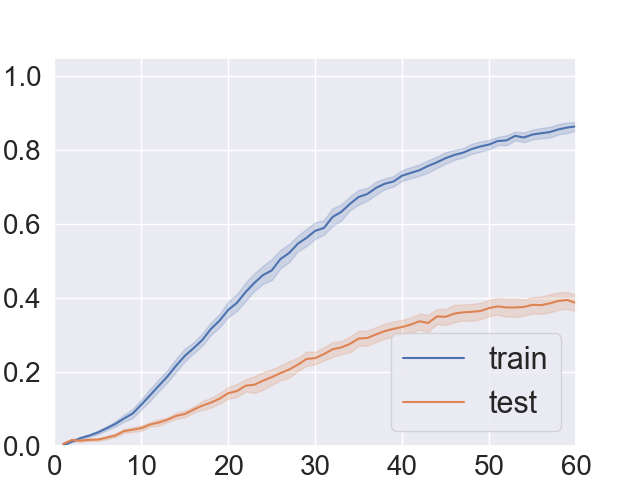}
  \caption{Lst Acc flex+op}
  \label{fig:flex_lst_sv_acc} 
\end{subfigure}

\begin{subfigure}{.25\textwidth}
  \centering
  \includegraphics[width=\columnwidth]{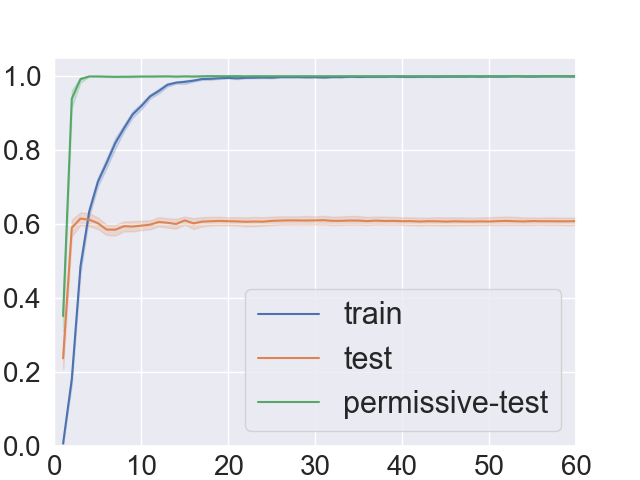}
  \caption{Spk Acc fix+op}
  \label{fig:fix_spk_sv_acc}
\end{subfigure}%
\begin{subfigure}{.25\textwidth}
  \centering
  \includegraphics[width=\columnwidth]{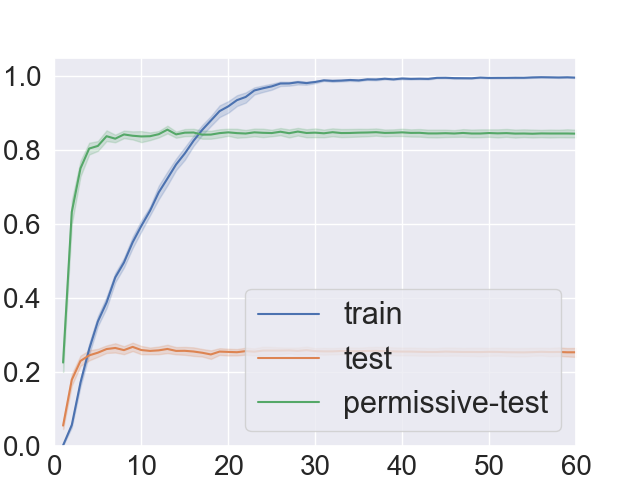}
  \caption{Spk Acc flex+op}
  \label{fig:flex_spk_sv_acc} 
\end{subfigure}

\begin{subfigure}{.25\textwidth}
  \centering
  \includegraphics[width=\columnwidth]{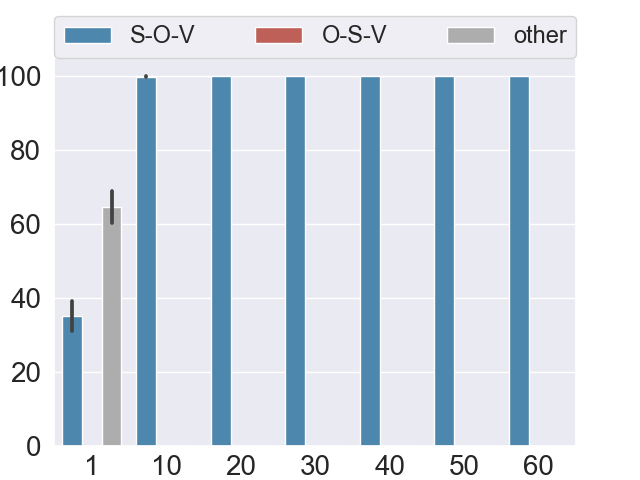}
  \caption{\%Order fix+op}
  \label{fig:fix_spk_sv_order}
\end{subfigure}%
\begin{subfigure}{.25\textwidth}
  \centering
  \includegraphics[width=\columnwidth]{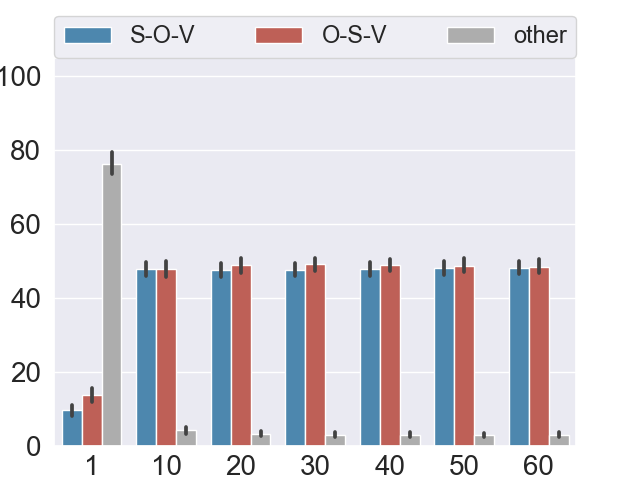}
  \caption{\%Order flex+op}
  \label{fig:flex_spk_sv_order}
\end{subfigure}

\begin{subfigure}{.25\textwidth}
  \centering
  \includegraphics[width=\columnwidth]{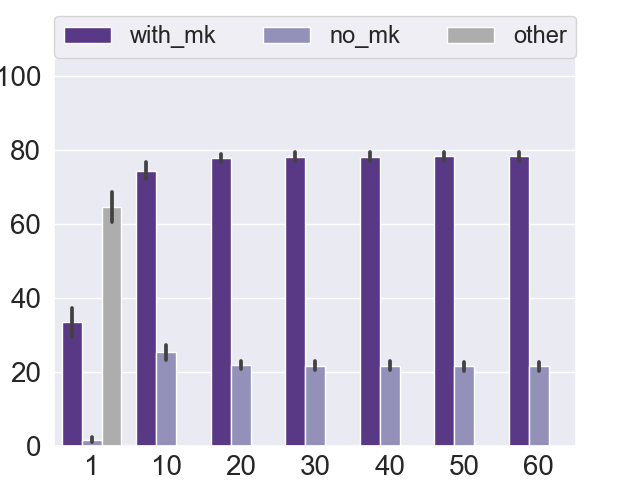}
  \caption{\%Case fix+op}
  \label{fig:fix_spk_sv_mk}
\end{subfigure}%
\begin{subfigure}{.25\textwidth}
  \centering
  \includegraphics[width=\columnwidth]{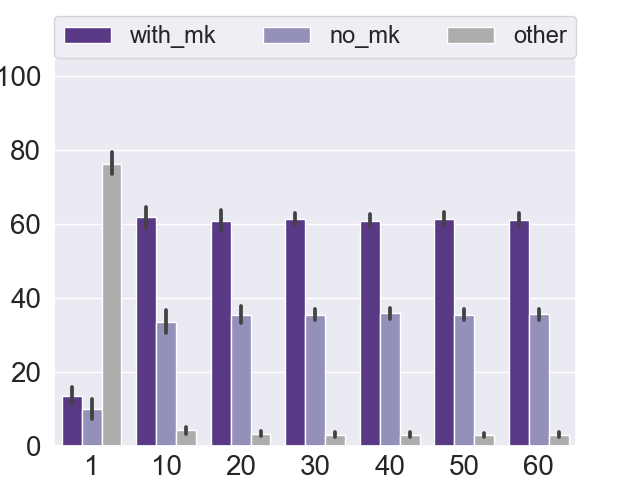}
  \caption{\%Case flex+op}
  \label{fig:flex_spk_sv_mk}
\end{subfigure}
\caption{Supervised learning results across training epochs for the fixed- (left) and flexible-order (right) language: accuracy of listening (a,b) and 
speaking (c,d) agents;
distribution of word order (e,f) and markers (g,h) in speaker-generated utterances.
All results are averaged over 20 random seeds.
}
\label{fig:sv}
\end{figure}

\subsection{Accuracy}
\label{sec:accSL}
During evaluation, both types of agents generate their predictions by greedy decoding.
Accuracy is computed at the whole utterance or meaning level. Specifically, \textbf{listening accuracy} is 1 if all of $A$, $a$, and $p$ are correct, otherwise it is 0.
Speaking accuracy is evaluated in two ways:
(i)~Regular \textbf{speaking accuracy} is 1 only if the generated utterance is identical to the one in the dataset. 
(ii)~\textbf{`Permissive' speaking accuracy}
considers the fact that our grammars admit multiple utterances for the same meaning:
for each test sample, we generate all correct candidates (i.e., with or without marker; OSV and SOV for the flexible-order language). Permissive accuracy is 1 if the generated utterance matches any of the candidates. 
As long as the utterance is acceptable, matching an arbitrary choice of order or marking for a given meaning does not matter.
Hence, the discussion in this section is based on permissive speaking accuracy.
Fig.~\ref{fig:sv} shows accuracy results for both agent types, each averaged over 20 random initialization seeds. 

We find that  our agents learn to speak and  understand the fixed-order language with extremely high accuracies (Fig.~\ref{fig:fix_lst_sv_acc}, \ref{fig:fix_spk_sv_acc}). 
By contrast, the flexible-order language reaches only 
38.7\% listening accuracy (Fig.~\ref{fig:flex_lst_sv_acc}) and 84.5\% permissive speaking accuracy (Fig.~\ref{fig:flex_spk_sv_acc}) on average for the unseen test. Note this does not reflect a weakness of the learners, but the ambiguity of the language itself: namely, subject and object are not distinguishable when the marker is absent, which happens in a third of the utterances.\footnote{The ability of RNNs to learn fixed-order languages equally well as their flexible-order/case-marking has been demonstrated by previous studies \cite{lupyan2002case,chaabouni-etal-2019-word,bisazza-etal-2021-difficulty}, but only when case marking is consistently present.}
These results are consistent with the higher comprehension and production accuracy of human participants learning the fix+op vs. flex+op language in \citet{fedzechkina2017balancing}.
Specifically, their flex+op group reached 96\% comprehension accuracy with 6.2\% grammatical mistakes, while the fix+op group reached 99\% accuracy with no grammatical mistakes (see Section 3.1 in \citet{fedzechkina2017balancing}).
Next, we inspect the properties of the language generated by speaking agents during the learning process.

\subsection{Production Preferences}
\label{sec:SV_production}
Fig.~\ref{fig:fix_spk_sv_order}, \ref{fig:flex_spk_sv_order} show the proportion of SOV vs. OSV test utterances generated by the speaking agents across training epochs (see Appendix~\ref{append:proportion} for details on utterance categorization).
For both languages, learners show a clear \textbf{probability-matching behavior}: in a few epochs, the order distribution becomes the same as in the input language and remains unchanged throughout the whole training.  
A similar pattern is visible for marking (Fig.~\ref{fig:fix_spk_sv_mk}, \ref{fig:flex_spk_sv_mk}).
Looking closer at fix+op (Fig.~\ref{fig:fix_spk_sv_mk}) we notice a slightly higher production of cases than the initial 66.7\%, which is even less efficient than the input language.

\SHORTENED{\gap}

Taken together, these results show that our agents are good learners but do not regularize the use of the two strategies in a human-like way after SL, which is in line with the iterated supervised learning results of \citet{chaabouni-etal-2019-word} and \citet{lian-etal-2021-effect}. 
This leads us to the next phase: optimizing agents for communicative success.

\section{Communication Learning Results}
\label{sec:rl_result}
We study the effect of communication learning on communication success and language properties. % agents' communication by looking at: 

\subsection{Communication Success}
\label{sec:rl_result_success}

Once a pair of agents is trained to speak/listen, 
they start communicating with each other to achieve a shared goal: the listener should understand the speaker, i.e. reconstruct the intended meaning. %the listener needs to successfully reconstruct the speaker's input meaning according to its produced message. 
Task success is evaluated by \textbf{meaning reconstruction accuracy}, which corresponds to the listening accuracy (Section.~\ref{sec:accSL}) of a listener receiving a speaker-generated utterance as input.\footnote{Greedy decoding is used for both speaker and listener during the evaluation of communication success.}

The results in Fig.~\ref{fig:fix+op_Comm_Acc}, \ref{fig:flex+op_Comm_Acc} show that agents understand each other better after several communication  rounds. % and achieve a relatively high communication success in both languages, according to the increasing task success rate.
More specifically, the non-ambiguous language (Fig.~\ref{fig:fix+op_Comm_Acc}) suffers from an initial drop but recovers the initial accuracy by epoch 20.
The ambiguous language (Fig.~\ref{fig:flex+op_Comm_Acc}) starts from a lower communication success rate as expected %(cf. Section.~\ref{sec:accSL})
but becomes more and more informative throughout communication.
In particular, around epoch 40, agents recover the communication success they had achieved at the end of SL on known meanings (85.2\%) while even  \textit{exceeding} it for new meanings (61.5\% vs. 38.7\%).
These results strongly suggest the language becomes less ambiguous by interaction.

Additionally, we report a noticeable drop in average performance towards the last epochs. The individual seed results reveal that most agent pairs suffer from a collapse of their communication protocol in the final stages of RL.
We attribute this issue to a known limitation of the REINFORCE algorithm related to its high gradient variance \cite{lu2020countering}.
Having assessed that our NeLLCom agents are able to learn a language and use it for conveying meanings, we now inspect \textit{how} their language changes during communication.

\subsection{Production Preferences}
\label{sec:comm-prod-pref}

The proportions of word order and case markers generated by the speaking agents are shown respectively in Fig.~\ref{fig:fix+op_Order}, \ref{fig:flex+op_Order} and Fig.~\ref{fig:fix+op_mk}, \ref{fig:flex+op_mk} (see Appendix~\ref{append:proportion} for details on utterance categorization).
We can see that these properties change considerably during communication learning, which was not the case during SL.
The increase of communication success observed in both languages already indicates that languages tend to become more informative. 
The key question is whether informativity is being balanced with efficiency, in a similar way as observed in human experiments \cite{fedzechkina2017balancing}.

\begin{figure}[!ht]
\centering
\begin{subfigure}{.25\textwidth}
  \centering
  \includegraphics[width=\columnwidth]{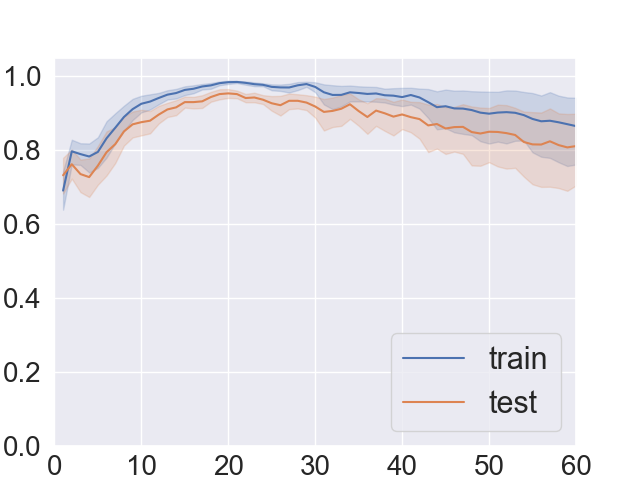}
  \caption{Comm.acc. fix+op}
  \label{fig:fix+op_Comm_Acc}
\end{subfigure}%
\begin{subfigure}{.25\textwidth}
  \centering
  \includegraphics[width=\columnwidth]{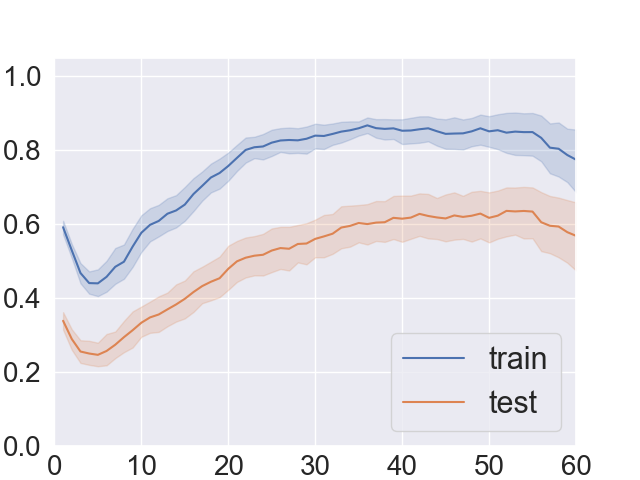}
  \caption{Comm.acc. flex+op}
  \label{fig:flex+op_Comm_Acc}
\end{subfigure}

\begin{subfigure}{.25\textwidth}
  \centering
  \includegraphics[width=\columnwidth]{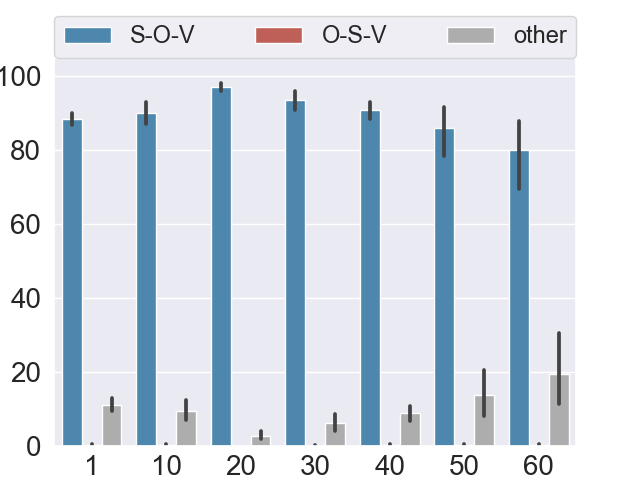}
  \caption{\%Order fix+op} 
  \label{fig:fix+op_Order}
\end{subfigure}%
\begin{subfigure}{.25\textwidth}
  \centering
  \includegraphics[width=\columnwidth]{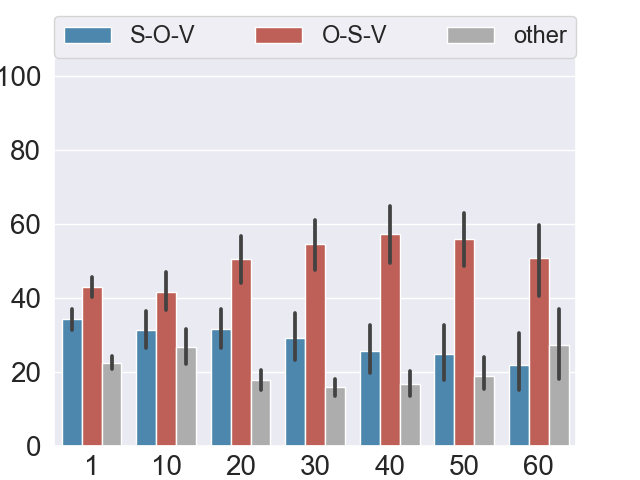}
  \caption{\%Order flex+op}
  \label{fig:flex+op_Order}
\end{subfigure}

\begin{subfigure}{.25\textwidth}
  \centering
  \includegraphics[width=\columnwidth]{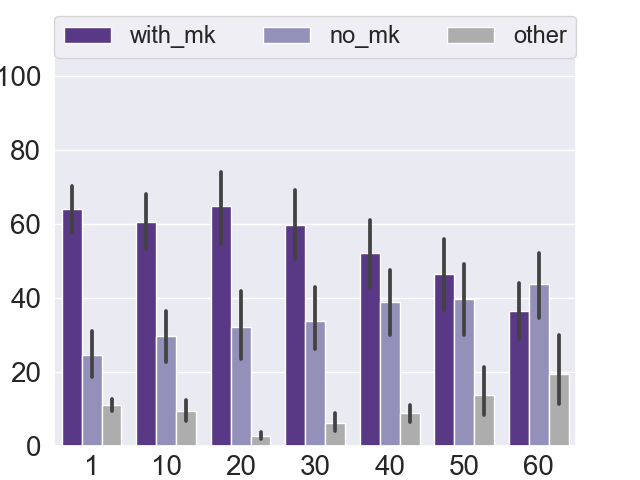}
  \caption{\%Marking fix+op} 
  \label{fig:fix+op_mk}
\end{subfigure}%
\begin{subfigure}{.25\textwidth}
  \centering
  \includegraphics[width=\columnwidth]{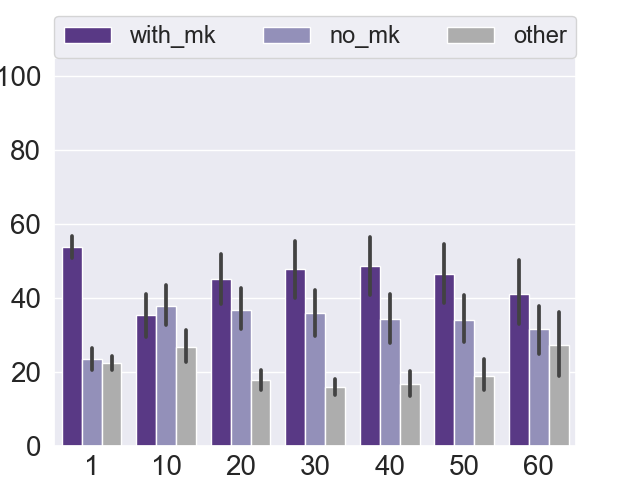}
  \caption{\%Marking flex+op}
  \label{fig:flex+op_mk}
\end{subfigure}%

\begin{subfigure}{.25\textwidth}
  \centering
  \includegraphics[width=\columnwidth]{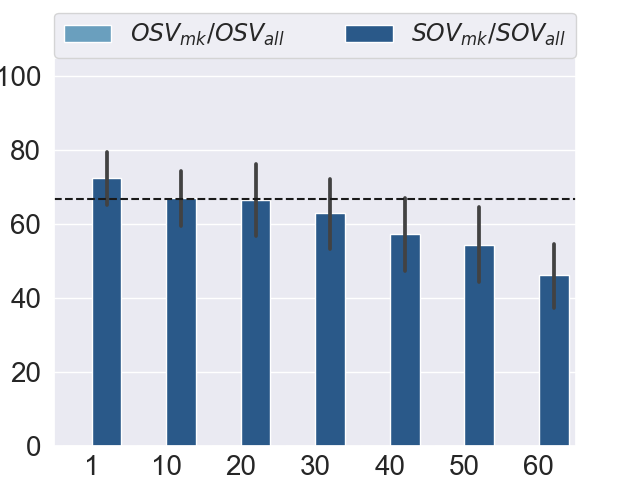}
  \caption{Cond.mark fix+op}
  \label{fig:mk_on_fix+op}
\end{subfigure}%
\begin{subfigure}{.25\textwidth}
  \centering
  \includegraphics[width=\columnwidth]{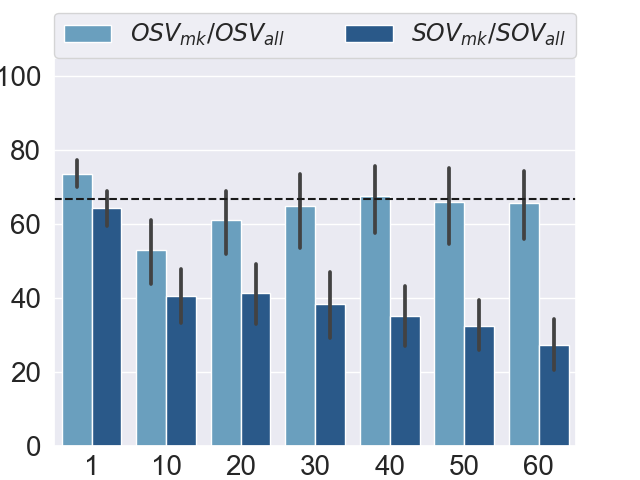}
  \caption{Cond.mark flex+op}
  \label{fig:mk_on_flex+op}
\end{subfigure}
\caption{
Communication learning results across training epochs for the fixed- (left) and flexible-order (right) language:
meaning reconstruction accuracy (a,b);
distribution of order (c,d) and markers (e,f) in speaker-generated utterances; marking conditioned on different orders (g,h).
Dashed lines indicate marking in the initial dataset (66.7\%).
All results averaged over 20 random seeds.
}
\label{fig:rl}
\end{figure}

\paragraph{Fix+op}
\label{sec:fix+op}
This language is redundant as it uses both fixed order (SOV) and marking to convey argument roles. 
As shown in Fig.~\ref{fig:fix+op_Order}, agents keep using SOV throughout the communication process.\footnote{The slight drop of SOV in the last epochs is due to the increase of non-classifiable (other) utterances, in turn related to the final communication collapse mentioned in Section.~\ref{sec:rl_result_success}.}
Similarly, human experiments of language emergence have shown that participants hardly ever create innovations in languages that are already systematic \cite{st2009relationships,tily2011learnability, fedzechkina2017balancing}. 
Importantly, Fig.~\ref{fig:fix+op_mk} reveals a clear preference towards dropping markers, as evidenced by a steady increase of \textit{no\_mk} utterances (light color). 
This aligns with the finding in \citet{fedzechkina2017balancing}, whereby human learners of the fixed-order language significantly reduced the use of marking over three days of training.\footnote{For detailed case marking results in human production, see Section 3.3 and Fig.4 in \citet{fedzechkina2017balancing}.}
The tendency to drop case markers is often explained by a human preference for reducing redundancy and increasing efficiency. Notably, the agents in our framework did not have any manually coded efficiency bias. The maximum allowed message length was much longer than the utterances needed to get the message across and the agents were not incentivized in any way to produce shorter sentences. 
Thus, we explain the observed pattern as a tendency of the neural agents to make the language more systematic as long as this does not harm communicative success.

\paragraph{Flex+op}
\label{sec:free+op}
Recall this language is originally as efficient as fix+op (i.e. same average utterance length) but less informative due to the presence of ambiguous utterances. 
We can think of at least two ways in which human or human-like learners could improve it, namely:
(i) keep using both orders interchangeably but use markers more systematically, or
(ii) choose one order as dominant and keep using markers optionally (or not at all).
Note that different pairs of speaking/listening agents may opt for different, though equally optimal strategies.

We find that NeLLCom agents increasingly produce OSV utterances (Fig.~\ref{fig:flex+op_Order}), reaching a situation where OSV is twice as common as SOV when communication success is at its highest (epoch $\sim$50).
At the same time, marker use fluctuates initially and then stabilizes around 55\%, that is still the majority of cases but less than the initial rate (66.7\%).
This strongly suggests that agents are making the language more informative while reducing effort, according to strategy (ii).
These results do not fully match those of \citet{fedzechkina2017balancing}, where most subjects instead adopted strategy (i).\footnote{For detailed word order results in human production, see Section 3.2 and Fig.3 in \citet{fedzechkina2017balancing}.}
Nonetheless, our findings provide important evidence that the word order/case marking trade-off can emerge in neural learners without hard-coded biases.

%\subsubsection{Case Marking conditioned on Orders}
\paragraph{Conditional case marking}
Besides \textit{how many} markers are used, it is important to understand \textit{how} they are used. As \citet{fedzechkina2017balancing} point out, learners of a flexible-order language could reduce uncertainty by conditioning their marker use on word order (asymmetric case marking).
For instance, using object marking only in SOV utterances could minimize uncertainty while maximizing efficiency. 
As discussed above, NeLLCom agents using flex+op tend to prefer an order over the other, however they are far from using one exclusively. Could our agents also be using markers conditionally?
Fig.~\ref{fig:mk_on_flex+op} shows the proportion of OSV utterances having a marker out of all OSV's (OSV$_{mk}$/OSV$_{all}$) and the same for SOV's.% (SOV$_{mk}$/SOV$_{all}$).%
\footnote{For completeness, Fig.~\ref{fig:mk_on_fix+op} shows conditional case marking results for fixed+op, however this is less interesting as word order is a sufficient disambiguation cue in this language.}
Indeed, agents use marking decreasingly when producing SOV utterances but maintain the marker percentage in OSV utterances, which matches unexpectedly well the human tendency observed by \citet{fedzechkina2017balancing}, Section 3.4.
Whether this is due to a coincidence or to a bias (e.g. towards marking the first entity appearing in an utterance) remains for now unexplained.

\begin{figure*}[!ht]
\centering
  \includegraphics[width=\textwidth]{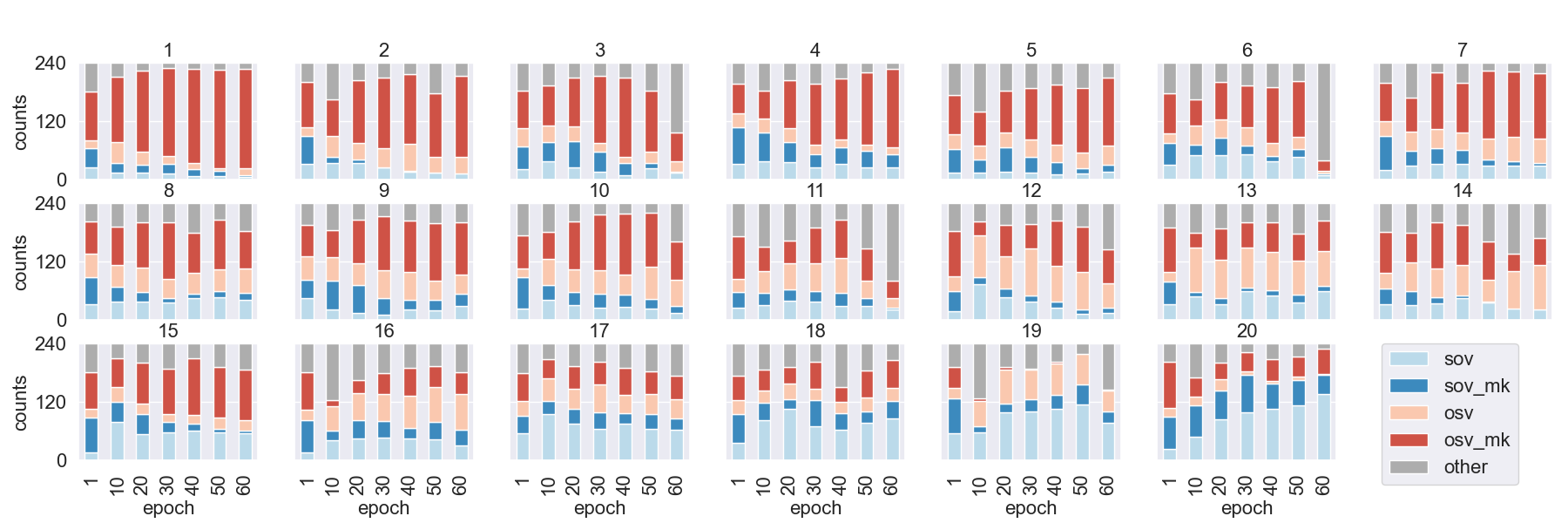}
\caption{Individual production distributions (flex+op language). Utterances are categorized into 5 types, namely SOV without marker, SOV with marker, OSV without marker, OSV with marker and uncategorized (other). Color denotes word order (blue: SOV, red: OSV), shading denotes marking (dark: with marker, light: without). 
Subplots are manually arranged to highlight clusters of similar trajectories.
}
\label{fig:distribution}
\end{figure*}

\begin{figure*}[!ht]
\centering
\begin{subfigure}{.34\textwidth}
  \centering
  \includegraphics[width=\columnwidth]{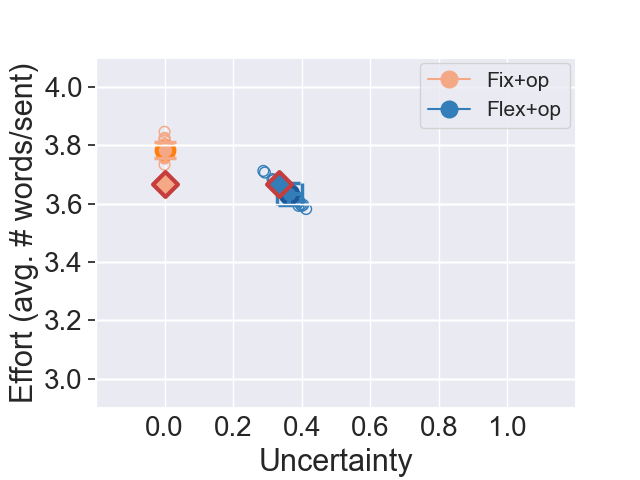}
  \caption{Supervision Results}
  \label{fig:diamond_sv}
\end{subfigure}%
\begin{subfigure}{.34\textwidth}
  \centering
  \includegraphics[width=\columnwidth]{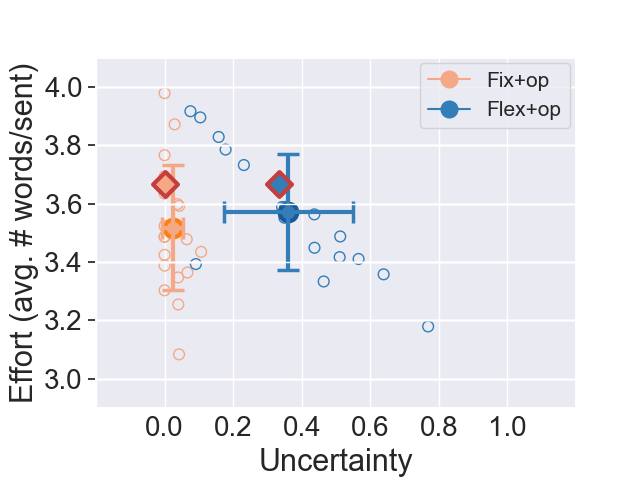}
  \caption{Communication Results}
  \label{fig:diamond_rl} 
\end{subfigure}%
\begin{subfigure}{.3\textwidth}
  \centering
  \includegraphics[width=\columnwidth]{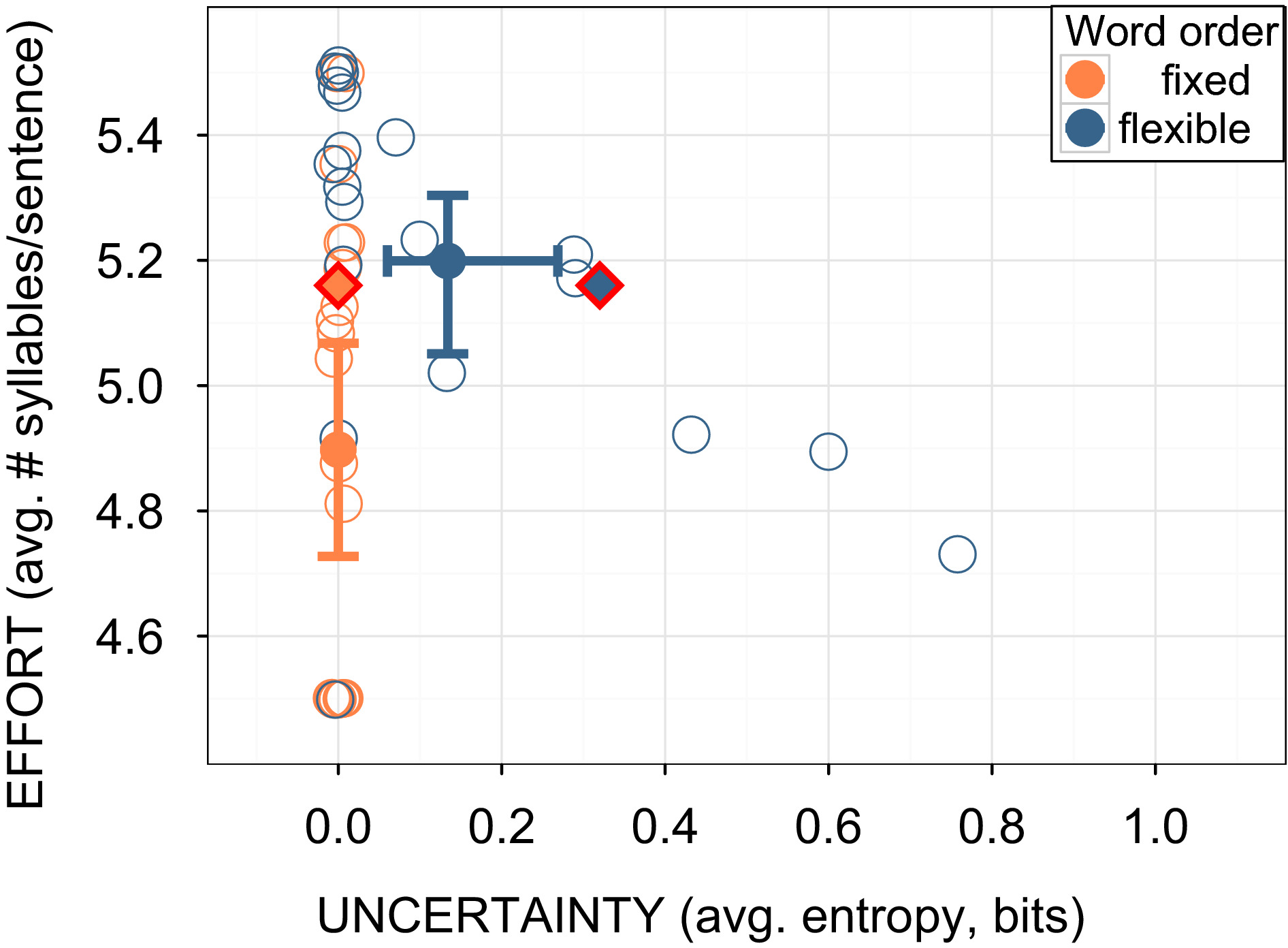}
  \caption{Human Results}
  \label{fig:diamond_human}
\end{subfigure}

\caption{Uncertainty ($H$) versus production effort: 
NellCom agents' results after supervised (a) and communication learning (b);
human results on last day of training (c), reproduced with permission from \citet{fedzechkina2017balancing}.
Solid diamonds mark the initial uncertainty-effort value for each language.
Empty circles represent the individual 20 agent pairs. 
Solid circles are the average of all agent pairs. 
}
\label{fig:diamond}
\end{figure*}

\SHORTENED{
\section{Further Analysis}
\label{sec:ana}
Section.~\ref{sec:rl_result} analyzed communication results averaged over multiple randomly initialized agents. 
Here, we look at possible variations among pairs of speaking-listening agents 
and study whether such variations can be explained by a single underlying principle.   
}

\section{Individual Learners' Trajectories}
\label{sec:individuals}

All results so far were averaged over multiple randomly initialized agents. 
Here, we look at possible variations among pairs of speaking-listening agents. 
%and study whether such variations can be explained by a single underlying principle. 
We focus only on the flexible-order language, as it is more likely to undergo different optimization strategies. % (cf. Section.~\ref{sec:comm-prod-pref}). \AB{right Yuchen?}
Fig.~\ref{fig:distribution} shows 20 production distributions, each corresponding to a different random seed.
Most agents (no.~1 to 14) regularize their productions towards the OSV order, as anticipated by the average results in Fig.~\ref{fig:rl}.
However, we also find two agent pairs that take the opposite path and produce more SOV (no.~19 and 20).
The remaining four agents show no clear order preferences (no.~15, 16, 17 and 18).
As for case marking, a clear preference to drop the marker from SOV utterances can be found in 15/20 pairs (no.~4, 5, 9, 10 16 are exceptions), which reflects the average trend of conditional case marking shown in Fig.~\ref{fig:mk_on_flex+op}.
This high degree of between-agents variability matches human results \citep{fedzechkina2012language, fedzechkina2017balancing, culbertson2012learning, hudson2005regularizing} where learners often adopt different strategies to reach a common optimization objective.

\SHORTENED{
\subsection{Uncertainty/Efficiency Trade-off}}
\paragraph{Uncertainty/efficiency trade-off}

We explore whether the observed trajectories can be explained by a single principle: a trade-off between uncertainty and efficiency.
Following \citet{fedzechkina2017balancing},  we quantify \textbf{production effort} as the average number of words per generated utterance.\footnote{\citet{fedzechkina2017balancing} used the  number of syllables, but that correlated perfectly with the number of words.}
To quantify \textbf{uncertainty}, we use their ``conditional entropy over grammatical function assignment'' ($H$), which captures the uncertainty over the intended meaning experienced by a listener with  perfect knowledge of the initial grammar (see detailed definition in Appendix.~\ref{append:h}).
Fig.~\ref{fig:diamond} presents uncertainty versus production effort at three time points: the initial language defined by the grammar, production after SL, and production after communication.
For comparison, the human results  of \citet{fedzechkina2017balancing} are reported in Fig.~\ref{fig:diamond_human}.

In Fig.~\ref{fig:diamond_sv}, the tight distribution of data points (empty circles) around the initial state (diamonds) reconfirms that SL alone does not lead to meaningful regularization. In fact, the only noticeable drift happens for fix+op in the counter-intuitive direction of increasing effort in the absence of uncertainty, as also anticipated by Fig.~\ref{fig:fix_spk_sv_mk}.
\SHORTENED{}
Communication results (Fig.~\ref{fig:diamond_rl}) show a very different picture: for both languages, average effort appears to \textit{decrease} without noticeably increasing uncertainty.
%with an unnoticeable increase of uncertainty.
Variability among agents is also wide, as already noticed in the qualitative analysis of Section.~\ref{sec:individuals}.
In fix+op, 17/20 agents produce shorter sentences.
\citet{fedzechkina2017balancing} report effort reductions in 14/20 participants.
In flex+op, the average uncertainty/effort values do not deviate much from the initial state, but individual data points reveal an unmistakable pattern, namely an inverse linear correlation between effort and uncertainty (empty blue circles in Fig.~\ref{fig:diamond_rl}).
We closely inspect three instances:
(i) The top-left data point (H=0.08, E=3.91) corresponds to agent pair no.~1 in Fig.~\ref{fig:distribution} whose language becomes fixed-order (OSV) and fully marked, i.e. unambiguous but inefficient. 
(ii) The bottom-right data point (H=0.77, E=3.19) corresponds to no~.19 in Fig.~\ref{fig:distribution} where most markers are dropped (5\% for OSV and 24\% for SOV) but no order strongly dominates, resulting in high ambiguity. 
(iii) Finally, the data point at (U=0.09, E=3.39) represents the only clear outlier from the linear correlation. This agent pair, corresponding to no.~20 in Fig.~\ref{fig:distribution}, succeeds at minimizing \textit{both} effort and uncertainty by using SOV predominantly (76\%) and reserving most markers to the less common order OSV (highly asymmetric case marking).
Interestingly, no outliers are found on the other side of the line: i.e. none of the 20 agents pairs appears to increase both effort and uncertainty, just like in the human results (Fig.~\ref{fig:diamond_human}).

\section{Discussion and Conclusion}

We studied the conditions in which the word order/case marking trade-off, a well established language universal example, could emerge in a small population of neural-network learners.
We hypothesized that more naturalistic settings of language learning and use could lead to more human-like results,
without the need to hard-code specific biases, such as least effort, into the agents.
We then proposed a new Neural-agent Language Learning and Communication framework (NeLLCom) where pairs of speaking and listening agents learn a given language through supervised learning, and then use it to communicate with each other, optimizing a shared reward via reinforcement learning.

We used NeLLCom to replicate the experiments of \citet{fedzechkina2017balancing}, where two groups of human participants were asked to learn a fixed- and a flexible-order miniature language, respectively, and to use it productively after training. 
Our results with RNN-based meaning-to-sequence and sequence-to-meaning networks confirm that SL is sufficient for perfectly learning the languages, but does not lead to any human-like regularization, in line with recent simulations of iterated learning \cite{chaabouni-etal-2019-word,lian-etal-2021-effect}.
By contrast, communication learning leads agents to modify their production in interesting ways:
Firstly, optional markers are dropped more frequently in the redundant fixed-order language than in the ambiguous flexible-order language, which matches human learning results.
Moreover, one of the two equally probable word orders in the flexible-order language becomes clearly dominant and case marking starts to be used consistently more often in combination with one order than with the other. This conditional use of marking also matches human results.
Some interesting differences were also observed: for instance, NeLLCom agents showed, on average, a slightly stronger tendency to reduce effort rather than uncertainty. 
As another difference, several human subjects managed to `break' the linear correlation by making the language more efficient \textit{and} less uncertain, whereas this happened only in one of our agent pairs.
Despite these differences, %the overall trends in our 
agents' productions 
show a clear correlation between effort and uncertainty, which strongly matches the core finding of
%remain similar to those reported by 
\citet{fedzechkina2017balancing}.
We conclude that the word order/case marking trade-off as a specific realization of the efficiency/informativity trade-off can, in fact, emerge in neural network learners equipped with a need to be understood.

We made an important step towards developing a neural-agent framework that replicates patterns of human language change without the need to hard-code ad-hoc biases.
Future work includes extending the current framework with iterated learning, which might lead agents to further optimize the ambiguous language and improve communication success over generations. We also plan to experiment with different neural network architectures to study the impact of architecture-specific structural biases, and with different word order universals.

We hope our framework will facilitate future simulations of language evolution at different timescales with the end goal of explaining why human languages look the way they do.

\iftaclpubformat

\section*{Acknowledgments}
Lian was partly funded by the China Scholarship Council (CSC 201906280463). 
Bisazza was partly funded by the Dutch Research Council (NWO project InDeep NWA.1292.19.399).
We thank the editors and reviewers for their helpful comments. 
\else
\fi

\iftaclpubformat

\bibliography{tacl2021,anthology}
\bibliographystyle{acl_natbib}

% \section{Appendix}

\appendix
\section{Datasets and Model training} 
\label{app:trainDetails}
\paragraph{Datasets}

Each word in a language corresponds uniquely to an entity or an action in the meaning space, leading to vocabulary size $|V|$= 8+10+1(marker) =19.
After the train/test split, we check that each entity and each action in the test set appears at least once in the training.
If that's not the case, we randomly swap the meaning-utterance pairs containing unseen entities with random ones from the training set.
An end-of-sentence $\langle$\textsc{eos}$\rangle$ token is appended to each utterance and 
padding is used to deal with variable utterance lengths. 

\paragraph{Model training}
Hyper-parameters were set in preliminary SL experiments: 
Speakers have 8-dim. embeddings and a 128-dim. GRU layer. Listeners have 32-dim. embeddings and a 32-dim. GRU layer. 
A default Adam optimizer \cite{kingma2014adam} in PyTorch \cite{paszke2017automatic} is used for both SL and RL, with learning rate 0.01 and batch size 32.
Each training phase lasts 60 epochs 
and we repeat each experiment with 20 different random seeds.

\section{Utterance Length and Production Preferences}
\label{append:proportion}
In principle, RNN can generate sequences of variable length. In practice, this is achieved by fixing a maximum message length (10 words in our setup) and truncating the sequence when the first symbol $\langle$\textsc{eos}$\rangle$ is generated.
We noticed, however, that during communication our speaking agents do not always end their message with $\langle$\textsc{eos}$\rangle$, but rather duplicate their final words to fill up the maximum utterance length after generating a well-formed initial message. 
As long as the first part of the utterance perfectly matches one of the structures admitted by the grammar,
we truncate the utterance at the last word before duplication. On average, this affects 15\% of the utterances by epoch 60.

Speaker-generated utterances for the unseen meanings (240 in total) are then classified into five types: 
SOV without marker, SOV with marker, OSV without marker, OSV with marker 
and uncategorized (other). Properties are computed as: 

\vspace{1mm}
\begin{tabular}{l}
\small$\%\mathrm{SOV} = (\mathrm{SOV_{mk}} + \mathrm{SOV_{no\_mk}}) / \mathrm{Total}$ \\
\small$\%\mathrm{OSV} = (\mathrm{OSV_{mk}} + \mathrm{OSV_{no\_mk}}) / \mathrm{Total}$ \\
\small$\%\mathrm{with\_mk} = (\mathrm{SOV_{mk}} + \mathrm{OSV_{mk}}) / \mathrm{Total}$ \\
\small$\%\mathrm{no\_mk} = (\mathrm{SOV_{no\_mk}} + \mathrm{OSV_{no\_mk}}) / \mathrm{Total}$
\end{tabular}

\section{Uncertainty Measure}
\label{append:h}
% \paragraph{Uncertainty Measure}
% \label{append:h}
This measure taken from \citet{fedzechkina2017balancing} captures the uncertainty about the role of the two entities expressed in an utterance, which is experienced by a listener with perfect knowledge of the grammar. 
It is formalized as the conditional entropy of grammatical function assignment (GF)  given sentence form (s.form):
\begin{multline}\small
H(\mathrm{GF|s.form}) = - \sum_{\mathrm{GFs}}\ \sum_{\mathrm{s.forms}}\\
p(\mathrm{s.form, GF}) * log_2p(\mathrm{GF|s.form})
\end{multline}

\noindent
According to the constraints of each grammar, possible sentence forms are $\mathrm{s.forms}$=$\{$SOV,OSV$\}$ and function assignments
$\mathrm{GFs} =\{ \mathrm{N_1 N_2 V,\ N_1 mk N_2 V,\ N_1 N_2 mk V}\}$.
Initial language uncertainties are as in \citet{fedzechkina2017balancing}: 0 for fix+op and 0.33 for flex+op.

\end{document}